\documentclass{article}
\usepackage{spconf,amsmath,graphicx}
\usepackage{latexsym}
\usepackage{soul}
\usepackage{color}
\usepackage{array}

\newcommand\mn[1]{\textcolor{black}{#1}}
\newcommand\mnm[1]{\textcolor{black}{#1}}

\newcommand\mt[1]{\textcolor{black}{#1}}
\newcommand{\md}[1]{\textcolor{black}{#1}}
\newcommand{\mmd}[1]{\textcolor{black}{#1}}

\title{One-to-Many Multilingual End-to-End Speech Translation}
\name{Mattia A. Di Gangi$^{1,2}$, Matteo Negri$^{1}$, Marco Turchi$^{1}$}
\address{$^1$Fondazione Bruno Kessler, MT unit, Italy \\
$^2$University of Trento, DiSI, Italy}
\begin{document}

\maketitle
\begin{abstract}

Nowadays, training end-to-end neural models for 
spoken language translation (SLT) still has  to confront with extreme data scarcity conditions. The existing SLT parallel corpora are indeed orders of magnitude smaller than those available for the closely related tasks of automatic speech recognition (ASR) and machine translation (MT), which usually comprise tens of millions of instances. To cope with data paucity,
in this paper we explore the effectiveness of transfer learning in end-to-end SLT by presenting a multilingual approach to the task. Multilingual solutions are widely studied in MT and usually rely on  ``\textit{target forcing}'', in which multilingual parallel data are combined to  train a single model by prepending to the input sequences a language token that specifies the target language.
However, when tested in speech translation, our experiments show that MT-like \textit{target forcing}, used as is, is not effective in discriminating among the target languages. Thus, we propose a variant that \md{uses target-language embeddings to shift the input representations in different portions of the space according to the language, so to }
better support the production of output in the desired
target language. Our experiments on \mnm{end-to-end} SLT from English into six languages show important improvements when translating into similar languages, especially when these are supported by scarce data. Further improvements are obtained when using English ASR data as an additional language (up to $+2.5$ BLEU points).
  
\end{abstract}


\section{Introduction}
The state-of-the art results obtained by encoder-decoder models \cite{sutskever2014sequence} with sequence-to-sequence learning in fields like ASR \cite{amodei2016deep,chan2016listen,chiu2017state,zeyer2018improved} and, most importantly, machine translation \cite{vaswani2017attention,bojar2018findings}, have led to the recent proposal of sequence-to-sequence learning for direct speech-to-text translation \cite{berard2016listen,bansal2017towards}, that is translating from audio without an intermediate output representation.
Unfortunately, end-to-end models require large amounts of parallel training data \cite{koehn2017six} that are not yet available for the SLT task. 
Indeed,  while state-of-the-art sequence-to-sequence systems for ASR and MT are respectively trained on thousands of hours of transcribed speech \cite{Chiu18}  and tens of millions of parallel sentences \cite{DBLP:journals/corr/abs-1803-05567}, the largest publicly available SLT corpus comprises
about 500 hours of translated speech and few others amount to less than 300 hours each \cite{mustc19}.

To overcome this limitation, several works have proposed approaches that exploit in different ways the wealth of ASR or MT data available.
Multitask learning or transfer learning are generally used 
to exploit ASR data  \cite{weiss2017sequence,bansal2018pre,berard2018end,anastasopoulos2018tied} with positive results,
and the improvements are more evident when less training data for SLT are available.
Other approaches to overcome the low-resource condition are data augmentation \cite{jia2018leveraging} and knowledge distillation \cite{liu2019end}. \md{However, } \cite{sperber2019attention} showed that current direct models are not data-efficient in leveraging non-SLT data, \md{and \mnm{that the classic} ``cascade'' approach (\mnm{i.e. a pipelined architecture integrating ASR and MT\footnote{Though effective, cascade SLT solutions have some drawbacks that end-to-end SLT aims to overcome in the long run. Besides the higher architectural complexity, these include: error propagation (ASR errors are hard to recover by the MT component), and larger inference latency.}) still performs better}}.

In this paper, we take advantage of the recent release of MuST-C \cite{mustc19}, which provides parallel SLT data for eight languages, in order to study whether multilingual data can be used to train systems with better translation quality than unidirectional (i.e. \textit{one-to-one}) systems. Multilinguality has been widely explored in neural MT  \cite{zoph2016multi,dong2015multi,luong2015multi,firat2016zero,firat2016multi,lu2018neural}, where it is now commonly performed using the \textit{target forcing} mechanism \cite{hatoward,johnson2017google}, which enables translation to many languages (\{\textit{one,many}\}\textit{-to-many}) without changing the underlying NMT architecture. The idea is to prepend the source sentence with a token representing the target language, and all the sentences are processed using the same shared encoder-decoder architecture. Although this approach has been proposed for RNN-based NMT, it works even better \cite{lakew2018comparison} when using the Transformer \cite{vaswani2017attention} architecture. 
Target forcing has also been applied to multilingual speech recognition \cite{toshniwal2018multilingual,zhou2018multilingual} showing to improve the transcription quality, although multilingual ASR shows to be better than its monolingual counterparts even when the language token is not provided. 

A single model with shared parameters is particularly appealing in  low-resource scenarios \cite{lakew2017multilingual,johnson2017google} as it performs a sort of transfer learning between language directions. 
However, compared to \textit{one-to-one} models,  the results of a multilingual model usually degrade in the language directions supported by more training data.
Taking advantage of the MuST-C corpus, in this study we focus on the \textit{one-to-many} scenario and investigate what groups of target languages favor transfer learning in SLT. To the best of our knowledge, this represents the  first study on the effectiveness of the multilingual approach to SLT.

Along this direction, we proceed incrementally by first showing the limitations of MT-like target forcing and then by proposing and evaluating our  SLT-oriented enhancements.
First, our initial experiments show that the target forcing approach as proposed in \cite{johnson2017google} compares poorly with the unidirectional baselines. By looking at the output, we observe that the system produces sentences whose words are coherently in one language, but  in many cases the chosen language is wrong. 
As the system is not able to learn the co-occurrence between the  embedding of the language token and the words in the target language, we then propose to give a stronger learning signal by modifying the input content using the language embedding.
This, in practice, is repeated along the time dimension so to be propagated through the whole input sequence \mmd{(rather than being one single vector among thousands of others)}. 

Our experiments show that, by using this variant, translating into similar languages, i.e. Germanic (German and Dutch) and western European Romance (French, Italian, Spanish and Portuguese) leads to better average results than those obtained by unidirectional systems. 
However, and to our view unsurprisingly due to the difficulty to transfer knowledge across distant languages, the same 
improvements are not observed when merging more languages. Indeed, using  the six languages together as a target yields improvements only for the lesser-resourced language direction, while combining all the eight languages covered by MuST-C leads to performance degradations in all the cases. 
In our final experiments, we also added English to the target languages of all the multilingual systems, which are then trained for translation and ASR.
This provides a slight but consistent improvement to all results. 
Overall, in all but the two target languages with more training data, our best multilingual models outperform \textit{one-to-one} models of comparable size by at least $0.4$ BLEU points. In particular, on the least represented language direction in MuST-C (i.e. English$\rightarrow$Portuguese, for which the corpus includes 385 hours of translated speech), the observed performance improvement over the \textit{one-to-one} competitor (up to  $+2.5$ BLEU points) indicates the feasibility of the proposed approach to operate in low-resource conditions.

\section{Direct Speech Translation}

Direct speech translation is defined as the problem of generating a sequence $\mathbf{Y}$ representing a text in a target language, given an input signal $\mathbf{X}$ representing a speech in the source language. A model for this task can be trained by optimizing the likelihood $L = \log P(\mathbf{Y} \vert \mathbf{X}; \mathbf{\theta})$, where $\mathbf{\theta}$ is the vector of model's parameters. The loss is usually computed in an autoregressive manner such as $L=\sum_{t=0}^T \log P(\mathbf{y}_t \vert \mathbf{y}_{<t}, \mathbf{X}; \mathbf{\theta})$. 

\noindent
Recent works on end-to-end SLT used deep learning recurrent models based on LSTMs \cite{hochreiter1997long} that differ mainly in the number of LSTM layers and the use of convolution in the encoder  \cite{weiss2017sequence,berard2018end,jia2018leveraging}. However, recurrent models are characterized by slow training \cite{gehring2017convolutional,digangi2018deep}
and they have been replaced in MT by Transformer \cite{vaswani2017attention}.  In our experiments we used Speech-Transformer \cite{dong2018speech}, an adaptation of Transformer to ASR, which in recent previous work  has shown to perform really well with little modification also on monolingual end-to-end SLT \cite{digangi2019adapting}.

\begin{figure}
    \centering
    \includegraphics[width=\columnwidth]{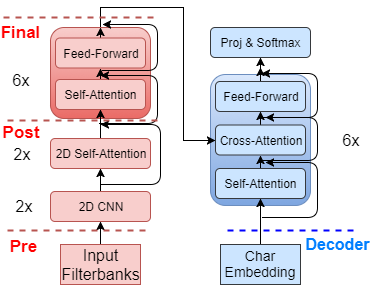}
    \caption{Proposed encoder-decoder architecture. The dashed lines represent the points where \textit{target forcing} can be applied.}
    \label{fig:method}
\end{figure}

\noindent
\textbf{Speech-Transformer} adapts Transformer to work with audio input provided as sequences of MEL filterbanks \cite{davis1980comparison}, which are characterized by joint dependencies in the two dimensions of time and frequency \cite{li2016exploring} and that are orders of magnitude  longer than the text representations handled by MT.
Because of those characteristics, the input is first processed and reduced with 2 layers of strided 2D CNNs \cite{lecun1998gradient}, each followed by ReLU activation and batch normalization \cite{ioffe2015batch}. 2D CNNs reduce the input dimension while capturing short-range bidimensional dependencies. The output of the second CNN layer is processed by two stacked 2D self-attention layers \cite{dong2018speech}, which are meant to model long-range bidimensional dependencies, then it is reshaped and processed by a feed-forward layer with ReLU activation. Then, the output of ReLU is summed to position vectors obtained through a trigonometric positional encoding \cite{vaswani2017attention} processed by a stack of Transformer encoder layers (darker box on the left in Figure \ref{fig:method}, which provides a simplified schema of the architecture). Each layer consists of a multi-head attention network for computing self attention of its input, followed by a 2-layered feed-forward network. Each of the two sublayers is followed by layer normalization \cite{ba2016layer} and residual connections \cite{he2016deep}. The decoder consists of a stack of Transformer layers, similar to the encoder layers except for the presence of an attention between encoder and decoder before the feed-forward network. The decoder receives in input a sequence of character embeddings summed with trigonometric positional encoding (right-hand side of the figure). 

\noindent
\textbf{2D Self-attention} (2DSAN) starts with three parallel 2D CNN layers that compute three different representation of their input $\mathbf{Q}, \mathbf{K}, \mathbf{V}$. Attention is computed as:

\begin{equation}
\label{eq:attn}
d_i = \textnormal{softmax}(\mathbf{Q}_i \mathbf{K}_i^T/\sqrt{d_{\textnormal{model}}}) \cdot \mathbf{V}_i    
\end{equation}

with $i = \{1, \dots, c\}$, where $c$ is the number of CNN filters, and the attention is computed in parallel for each $i$. Equation \ref{eq:attn} shows how attention along the time dimension is computed, but the three tensors are also transposed to compute attention along the frequency domain. All the attention outputs of the two parallel attention layers are then concatenated and processed by a final 2D CNN layer. All convolutions are followed by ReLU activation and batch normalization. This mechanism is analogous to the multi-head attention, except for the use of CNNs instead of affine transformations, and the use of the additional attention along the frequency domain.

\noindent
The authors of \cite{dong2018speech} mention the introduction of a distance penalty mechanism in their encoder self-attention layers, which is aimed to improve system's performance, but they do not provide additional details. Building on the same idea, we use a distance penalty that is subtracted to the input of softmax in Equation \ref{eq:attn}. Let $d = \vert i - j \vert$ be the distance between positions $i$ and $j$,  we then define a logarithmic penalty as follows:

\begin{equation}
    \pi_\textnormal{log} (d) = 
    \begin{cases}
    0\text{,}&\text{ if  d = 0} \\
    \log_e(d)\text{,}&\text{else}
    \end{cases}
\end{equation}

The use of logarithm is motivated by the fact that we want to bias the self attention towards the local context of each position, but we do not want to entirely prevent it from capturing the influence of global context. In preliminary experiments, we observed that using this logarithmic distance penalty mechanism leads to better results than an unbiased attention.

\section{Multilingual Translation}
In this section we first introduce the target forcing mechanism used in multilingual NMT and propose two variants to apply it to the SLT task. Then, we discuss two different but complementary ways to exploit ASR data in order to improve final translation performance.

\subsection{Target forcing}

We perform one-to-many multilingual speech translation with a single model by using the \textit{target forcing} mechanism \cite{johnson2017google}, which tags every source sentence with a language token indicating the target language.
In NMT, the language token is simply prepended to the input text sequence, and used identically to all the other tokens to retrieve an embedding that is jointly learned with the rest of the network. However, this method has to be adapted to a speech encoder that does not have an embedding layer. Thus, we propose two different approaches, namely: \textit{i)} \textit{concat}, which prepends a language embedding to the input sequence, and \textit{ii)} \textit{merge}, which sums the embedding to all the elements in the sequence. In both cases, the language embeddings are learnable parameters of the network.

\noindent
\textbf{Concat.} 
The first approach is a straightforward adaptation of the \textit{target forcing} mechanism \cite{johnson2017google} to speech translation. Let $\mathbf{X}\in R^{T\times F}$ be a matrix containing a sequence of audio feature vectors, where $T$ is the number of time steps and $F$ is the number of features, and let $\mathbf{l}\in R^{1xF}$ be a language embedding of the same size $F$. Then, analogously to the text input case, where the language token is prepended to the input string, we produce the new target-language dependent representation $\mathbf{X'}\in R^{(T+1)\times F}$ by concatenating $l$ to $X$. 
In this case, only the language embedding is a learnable parameter, while $\mathbf{X}$ is a fixed sequence of MEL filterbanks.

\noindent
\textbf{Merge.} 
While the \textit{concat} method modifies the input representation by concatenating an additional vector, the \textit{merge} approach alters the content of the input representation.
Given the two tensors $\mathbf{X}$ and $\mathbf{l}$ as in the previous case, now we define $\mathbf{X'}\in R^{TxF}$ as the sum $\mathbf{X'} = \mathbf{X} + \mathbf{l}$ where $\mathbf{l}$ is repeated along the time dimension. Considering the length of the audio input (recall that, differently from the textual representations handled by MT, the sequences of MEL filterbanks input to SLT are orders of magnitude  longer), the intuition is that the \textit{concat} approach can fail in propagating the language token towards the whole input sequence. In contrast,  with the \textit{merge} approach the 
\md{input tensor} is translated (in a geometric sense) to different portions of the space for each different language so to have representations that are clearly distinct between one language and another starting by the input. In this way, \md{the same input sentence has clearly different representations when it has to be translated to different language, and the learning task becomes easier.}

\subsection{ASR data}
It is widely demonstrated in literature that exploiting ASR data is useful for improving the performance of direct SLT systems, with the main approaches being transfer learning \cite{berard2018end,bansal2018pre} and multitask learning \cite{weiss2017sequence,anastasopoulos2018tied}. \md{Indeed, the ASR task is easier than translation as it is monolingual and does not involve reordering and, as such, it is used to obtain encoder representations that are better also for SLT.} In our experiments, we exploit ASR data in two ways. The first way is transfer learning, and consists in training another Speech Transformer for the ASR task, then using its encoder weights to initialize the parameters of the SLT model (pre-training). The second way is to use English ASR data as if it was an additional language for the multilingual system (+ASR). This approach is analogous to multi-task training but, unlike the approach used in \cite{weiss2017sequence} of using two different decoders, we can take advantage of the multilingual model and use ASR data to train also the decoder \cite{sperber2019attention}.
Because of its documented
effectiveness, we use pre-training in all our experiments, \md{while we evaluate the effectiveness of the +ASR approach with ablation experiments.}

\begin{table}[t!]
\centering
\begin{tabular}{l|c|c|c|c|c}
  \textbf{Tgt} & \textbf{\#Talk} &\textbf{ \#Sent} & \textbf{Hours}  & \textbf{src w} & \textbf{tgt w} \\ \hline
    \textbf{De} & 2,093 & 234K & 408  & 4.3M & 4.0M  \\\hline
    \textbf{Es} & 2,564 & 270K & 504  & 5.3M & 5.1M  \\\hline
    \textbf{Fr} & 2,510 & 280K & 492  & 5.2M & 5.4M  \\\hline
    \textbf{It} & 2,374 & 258K & 465  & 4.9M & 4.6M  \\\hline
    \textbf{Nl} & 2,267 & 253K & 442  & 4.7M & 4.3M  \\\hline
    \textbf{Pt} & 2,050 & 211K & 385  & 4.0M & 3.8M  \\\hline
    \textbf{Ro} & 2,216 & 240K & 432  & 4.6M & 4.3M  \\\hline
    \textbf{Ru} & 2,498 & 270K & 489  & 5.1M & 4.3M  \\\hline
\end{tabular}
\caption{Statistics for each section of MuST-C.}
\label{tab:stats}
\end{table}

\section{Experiments}
\textbf{Dataset.} The corpus we use is MuST-C \cite{mustc19}, which currently represents the largest publicly available multilingual corpus (one-to-many) for speech translation. It 
covers eight language directions, from English to German, Spanish, French, Italian, Dutch, Portuguese, Romanian and Russian. 
The corpus consists of audio, transcriptions and translations of English TED talks, and it comes with a predefined training, validation and test split.
In terms of hours of transcribed/translated speech (see Table \ref{tab:stats} for complete statistics), the size of the different sections of the corpus ranges from 385 (Portuguese) to 504 hours (Spanish).

\noindent
\textbf{Model settings.} The first two CNNs in the encoder have 16 output channels, $3\times 3$ kernel and stride $(2, 2)$. The CNNs inside the 2D self-attention have $3\times 3$ kernels, 4 output channels and stride $1$. The output CNN of the 2D self-attention has $16$ output channels. The following feed-forward layer has $512$ output features, which is the same size as the Transformer layers. The hidden feed-forward layer size of Transformer is $1024$. The decoder layers have also size $512$ and hidden size $1024$. Dropout is set to $0.1$ in each layer. 
Each minibatch includes up to 8 sentences for each language and we update the gradient every 16 iterations. 
All the models are trained with the Adam \cite{kingma2014adam} optimizer with an initial learning rate of $0.0003$, then $4000$ warmup steps during which it increases linearly up to a max value, and then decreases with the inverse square root of the number of steps \cite{vaswani2017attention}. As the batch size depends on the number of languages, the maximum learning rate is increased with this number. We searched the best learning rate testing on a held-out set, and we selected $0.01$ for the experiments with the Germanic and Romance languages, and 0.02 for the experiments with 6 or all languages. 
Our baselines are based on \md{the same architecture } but each of them is trained only on one language pair.
\md{As additional stronger baselines, we also train } cascade models \md{that} concatenate an ASR and an MT system, where the ASR consists of an attention model using the same architecture of our direct SLT systems and the MT is a Transformer Base architecture \cite{vaswani2017attention}. 
All the models are implemented in Pytorch \cite{paszke2017automatic} within the fairseq toolkit \cite{gehring2017convolutional}. 

\noindent
\textbf{Experimental Settings.} In our first experiments we train two multilingual models, one for Germanic languages (German, Dutch), and one for western European Romance languages (Spanish, French, Italian, Portuguese). Although Romanian is also a Romance language, we keep it out from this experiment because its slavic influences make it quite different from the other 4 languages.
Romanian and Russian are finally used in one last experiment.

\noindent
\textbf{Data preprocessing.} From each audio segment we compute the MEL filter-banks \cite{davis1980comparison} with 40 filters, using overlapping windows of 25 ms and step of 10 ms. The resulting spectrograms are normalized by subtracting the mean and dividing by the standard deviation. All the texts are tokenized and the punctuation is normalized. 
In our cascade models, ASR systems are trained without punctuation in output and with lowercased text split into characters, while MT systems receive lowercased English text without punctuation in input, and produce text that preserves the casing and with punctuation in the target language as output.
Both source and target are split in subwords with BPE segmentation \cite{sennrich2015neural} using $32,000$ joint merge rules in MT.
In the target side of the SLT systems, texts are split into characters and punctuation is kept in the target texts. We do not use BPE for SLT because in our preliminary experiments it performed poorly.

\begin{table}[]
\begin{tabular}{l|ll|llll}
              & De   & Nl   & Es   & Fr   & It   & Pt   \\
Baseline      & 17.3 & 18.8 & 20.8 & \textbf{26.9} & 16.8 & 20.1 \\\hline
C-Pre    & 14.0 & 11.6 & 13.0 & 16.3 & 10.7 & 14.5 \\
C-Post   & 12.0 & 13.8 & 12.3 & 18.0 & 9.3  & 14.6 \\
C-Final  & 14.5 & 12.1 & 13.6 & 16.7 & 10.2 & 16.2 \\\hline
M-Pre     & 17.6 & 19.5 &  20.5  & 26.2  & 17.2  & 22.3  \\
M-Post    & 17.1 & 19.2 & 20.5 & 26.2 & 17.4 & 22.3 \\
M-Final   & 17.4 & 18.8 & 20.4 & 26.7 & 17.2 & 22.2 \\
M-Dec. & 17.3 & 19.1 & 20.6 & 26.2 & 17.2 & 22.0 \\\hline
M-Pre + ASR    & \textbf{17.7} & \textbf{20.0} & \textbf{20.9} & 26.5  & \textbf{18.0} & \textbf{22.6} \\\hline

\end{tabular}

\caption{Results with  \textit{concat} (C-*) and \textit{merge} (M-*) \textit{target forcing} on $6$ languages. The baselines are \textit{one-to-one} systems. All the other results are computed with one multilingual system for En$\rightarrow${De,NL} and one for En$\rightarrow${Es,Fr,It,Pt}.} 

\label{tab:first-exps}
\end{table}

\section{Results}
\textbf{Concat vs Merge.} Our first experiment consists in comparing the baselines with the multilingual models 
based on
the \textit{target forcing} mechanism. 
The results presented in Table \ref{tab:first-exps} show that \textit{concat} target forcing (C-Pre) is much worse than the baselines. However, note that our baselines are stronger than the ones reported in \cite{mustc19}. By looking at the translations, we found that the cause of the degradation is that many sentences are acceptable translations, but in a wrong language. We first hypothesize that the processing performed in the layer preceding the encoder self-attentional stack loses the language information. Thus, we concatenate the language embedding to the representations of the \textit{Post} and \textit{Final} positions (see Fig. \ref{fig:method}), both after the 2DSANs. The new results, listed in Table \ref{tab:first-exps}, show small and non consistent variations, and are still worse than the baselines in all languages. 
Our second hypothesis is that the networks are not able to learn the joint probabilities of language embeddings and character sequences because of a combination of factors: 
character-level translations (instead of sub-words), very long source-side sequences and \md{the source sides in the corpus are highly overlapping} between languages. Thus, we
assume that our networks can learn to discriminate better among target languages by giving a stronger language signal.
For this reason, we introduce the \textit{merge}  target forcing that forces the network to generate target-language dependent encoder representations by translating them in different portions of the space according to the target language. The results in Table \ref{tab:first-exps} show that \textit{merge} target forcing (M-*) is definitely better than \textit{concat} for all the target languages, \md{and} also obtains performance that is on par with or better than the baselines. \md{M-Pre is the system that shows high results more consistently in all languages, and} the largest improvement is observed in En-Pt, with over $2.0$ BLEU points of gain, \md{followed by }$+0.7$ in En-Nl.
The BLEU score slightly degrades for Spanish by $0.2\sim0.4$, and for French by $0.2\sim0.7$. 
Besides the three different language embedding positions in the encoder, we also performed experiments by applying target forcing on the decoder, but they show slightly worse performance.  Then, for the following experiments we will continue only with the \textit{Pre} position that results in the best average performance on all the language directions.

When adding also ASR data to our training set (M-Pre + ASR) we observe small but consistent improvements in all languages. In this case, there are improvements over the baselines larger than $1$ BLEU points in $3$ out of $6$ target languages: Dutch, Italian and Portuguese. Moreover, the system does not degrade in En-Es, the direction with the largest dataset available, and it is only $0.4$ BLEU points below the baseline in En-Fr.
\md{These last experiments show that the advantage of training ASR data are visible even when the SLT models have been pre-trained on the ASR task.}

\begin{table}[]
\small
\begin{tabular}{l|llllllll}
              & De   & Nl   & Es   & Fr   & It   & Pt & Ro & Ru   \\\hline
   &      \multicolumn{8}{l}{\textbf{Baseline}}  \\
   \hline
   & 17.3 & 18.8 & 20.8 & 26.9 & 16.8 & 20.1 & 16.5  & 10.5 \\\hline
   & \multicolumn{8}{l}{\textbf{Multilingual}}        \\\hline
6  & 17.3 & 18.4 & 20.0 & 25.4 & 16.9 & 21.8 & - & - \\
8  & 16.5 & 17.8 & 18.9 & 24.5 & 16.2 & 20.8 & 15.9 & 9.8 \\ \hline
   & \multicolumn{8}{l}{\textbf{+ ASR}}\\\hline
6  & 17.4 & 19.2 & 19.7 & 26.0 & 17.2 & 21.8 & - & - \\
8  & 15.9 & 17.2 & 18.3 & 23.7 & 15.1 & 19.9 & 15.5 & 9.7 

\end{tabular}
\caption{Results for multilingual direct SLT systems with 6 and 8 target languages.}
\label{tab:second-exps}
\end{table}

\noindent
\textbf{Number of languages.} When training a system with all the $6$ target languages (De, Nl, Es, Fr, It, Pt) together, in Table \ref{tab:second-exps} we observe results on par with the baseline for German and Italian, slightly worse results on Dutch and Spanish (respectively, -$0.4$ and -$0.8$), a larger degradation for French ($-1.5$), but also a large improvement on Portuguese ($+1.7$), although smaller than using $4$ languages plus ASR. When adding ASR data to the $6$ languages, we observe improvements in most languages, and the new system is worse than the baseline only for Spanish and French, although the gap for French has been reduced to $-0.9$.
\md{However}, using all \md{the 8 target} languages leads to worse results and adding ASR data \md{contributes to worsen the performance in this case}. 
We think that the reason is related to the relatively low number of parameters of our models ($\sim 33$ millions), which reduce their capability to learn and discriminate between a larger number of languages.

\begin{table}[]
\begin{tabular}{l|ll|llll}
              & De   & Nl   & Es   & Fr   & It   & Pt   \\
Baseline      & 17.3 & 18.8 & 20.8 & 26.9 & 16.8 & 20.1 \\\hline
M-Pre + ASR    & 17.7 & 20.0 & 20.9 & 26.5  & 18.0 & 22.6 \\\hline
BL-Cascade    & 18.5 & 22.2 & 22.5 & 27.9 & 18.9 & 21.5 \\
M-Cascade & 18.6  & 22.0 & 22.1 & 27.3 & 18.5 & 22.8 \\
\end{tabular}
\caption{Comparison of the Baseline and the best multilingual system with the single language cascade (BL-Cascade) and the multilingual cascade (M-Cascade)}
\label{tab:cascade}
\end{table}

\noindent
\textbf{Comparison with cascade.} In Table \ref{tab:cascade} we compare the single and our best multilingual systems with two different cascade models: the BL-Cascade, where the  ASR system is concatenated with the single MT system and the M-Cascade, where the ASR is concatenated with a multilingual NMT model trained on all the 6 language pairs. As expected, our direct SLT baselines are significantly worse than the BL-cascade systems with differences that range from $-1.0$ for French to $-3.4$ for Dutch. Comparing the BL-Cascade with the M-Cascade systems, we observe not significant variation for the Germanic languages, but lower results in $3$ out of $4$ Romance languages ($-0.6$ for French), with the single improvement of $+1.3$ for Portuguese. Thus, multilingual MT generally affects negatively the performance, being beneficial only for the lowest resourced languages. 

A comparison between our best multilingual SLT (M-Pre + ASR) and the cascade systems shows that our system is able to reduce the gap between them. On Portuguese, it is on par with the multilingual cascade and $1.1$ point above the BL-Cascade. For Dutch, the target language with the largest initial gap, the difference is reduced to $2$ BLEU points, and for Italian it is only $0.9$ lower than the best cascade. All these results show that the multilingual SLT system is a valid solution to enhance the performance of the end-to-end speech translation system and is able to reduce the gap with cascade systems. 

\begin{table}[]
\begin{tabular}{l|ll|llll}
              & De   & Nl   & Es   & Fr   & It   & Pt   \\\hline
M-Pre         & 95.7 & 98.5 & 97.2 & 94.6  & 95.3 & 96.6  \\
M-Pre + ASR   & 96.1 & 98.7 & 97.9 & 95.3  & 95.4 & 95.2 \\
\end{tabular}
\caption{Percentage of sentences in the correct language computed with langdetect. }
\label{tab:perc}
\end{table}

\noindent
\textbf{Language analysis.} To better understand the behaviour of the multilingual system, we automatically detected, by using langdetect\footnote{https://github.com/shuyo/language-detection} \cite{nakatani2010langdetect},
the language of each sentence translated by our multilingual direct SLT systems. 
The results, listed in Table \ref{tab:perc}, show that our M-Pre systems do not translate in the correct language in all the cases, although the percentage is higher than 95\%.
Then, when using also ASR data, the percentage of correct language increases slightly in all languages except for Portuguese. However, the improvement in the correct language does not correlate with the improvement in BLEU score.
\mn{This suggests that} the improvement in BLEU score of M-PRE + ASR comes from better translations and not from more sentences translated in the correct language.

\section{Discussion}
The classic \textit{target forcing} mechanism, \mt{which pre-pends the language embedding in the input sequence and is widely used in MT, resulted to be not effective for the SLT task. Our variation, which sums the language embedding to the entire input representation, shows to be effective by imposing a sharper distinction between encoder representations for different target languages \mn{that allows us to reduce} the gap in performance \mn{with respect} to a stronger cascade model.}
The causes of this behaviour will be investigated in future work, but here \mt{it is important to remark the differences with multilingual works in MT and ASR to understand the complexity of the SLT scenario.}
In MT, translations are performed at BPE-level, which results in a shorter input sequence than character-level, but also limits the vocabulary selection for each target language. Moreover, character-level representations mean that most of the vocabulary is shared between the languages, which can be a source of confusion. Multilingual ASR shares with our task the mapping \mn{from} audio to text, with long sequences in the source side. However, the language is the same in the source and target sides and, as such, it is more difficult for the model to confuse the target language. Toshniwal et al. \cite{toshniwal2018multilingual} have shown that a Transformer model trained on $9$ Indian languages with different scripts outperforms monolingual baselines without any training trick, and further improvements are obtained with the \textit{target forcing} mechanism. This case is significantly different from our task, where without target forcing the system has a probability of $1/N$ to translate into the correct language, with $N$ being the number of languages. 
\mn{Although these differences make multilingual SLT a challenging scenario, the results achieved by our systems indicate that the lack of data can be overcome by training a single multilingual model on more languages.}

Our experiments have also shown, consistently with the results in MT \cite{johnson2017google}, that multilingual translation is particularly beneficial for the less resourced language pairs. However, in order to obtain the best results, it is needed to keep the number of target languages limited and train on similar target languages. We have also found beneficial to use ASR data together with the other translation data, and we are not aware of works showing similar results on an MT task.
The reason \mn{of the significant performance drops when the number of languages increases}
is probably related to the model capacity. Indeed, our models 
have $\sim30$ million parameters, while large MT multilingual models can have one order of magnitude more parameters. Unfortunately, due to the large GPU memory occupation of SLT models, in particular with Transformer, we could not perform experiments with larger models.

\section{Conclusions}
We \mn{explored} one-to-many multilingual 
speech translation as a method to increase the training data size for direct speech translation.
Since in our experimental conditions target forcing is scarcely effective to discriminate among target languages, we proposed a variation that overcomes the problem. Our results show that this approach can produce important improvements in the target languages with less data.
Adding ASR data to the training set allows the multilingual SLT system to outperform baseline unidirectional (i.e. one-to-one) systems
in 3 out of 6 target languages (the lesser-resourced ones) with an improvement larger than 1 BLEU score point, and to have comparable performance in the languages with more data. When adding all the MuST-C target languages in a single system, the performance degrades showing the difficulties of the multilingual SLT model to manage an increasing number of unrelated target languages.
In future work, we plan to extend our method to many-to-many multilingual SLT, and to investigate strategies that take into account the peculiarities of this task, as well as evaluating the impact of model capacity to manage more languages.

\section*{Acknowledgements}
This  work  is  part  of  a  project  financially  supported  by  an Amazon AWS ML Grant.

\bibliographystyle{IEEEbib}
\bibliography{strings,refs}

\end{document}